\documentclass[11pt]{article}
\usepackage{cmap}
\usepackage[margin=1in]{geometry}
\usepackage[T1]{fontenc}
\usepackage{lmodern}

\usepackage{natbib}
\usepackage{booktabs}
\usepackage{array}
\usepackage{amsmath,amssymb}
\usepackage{xcolor}
\usepackage{hyperref}
\hypersetup{colorlinks=true,linkcolor=blue,citecolor=blue,urlcolor=blue}
\usepackage{caption}
\usepackage{graphicx}
\usepackage{enumitem}
\usepackage{xurl}

\title{\textbf{Does the Outfit Change the Price?}\\
\large Context Halo in Vision-Language Model Valuation of an Identical Low-Cost Necklace}

\author{Brianne Lee, M.S., M.P.H. \\ \small Independent Researcher \\ \small \texttt{briannelee011@gmail.com}}
\date{October 2026}

\begin{document}
\maketitle

\begin{abstract}
A chunky gold-tone chain necklace bought online for \$2.43 was photographed four ways: worn with formal attire against a wood-panelled wall, worn at a party, worn with casual clothing in a recycling yard, and laid flat with no person present. We asked six frontier language models to estimate its price in 1{,}164 independent and sequential API sessions (4{,}604 logged rows) and compared estimates across settings. Among the four models that answered image prompts, the same necklace was priced 1.9 to 3.3 times higher in the formal photograph than in the flat-lay (all $p<0.01$). Against the yard photograph, the ratio ranged from 1.23 (Grok~4.5, not significant) to 3.57 (Kimi~K3), with Claude Fable~5 at 3.32. A text-only version of the task, in which the setting is described in words, produced formal-to-yard ratios of 2.2 to 3.9 in all six models, so the effect does not require pixels. Flat-lay estimates themselves were 7.8 to 18.7 times the receipt price, which means that part of any headline ``outfit multiplier'' computed against the receipt is a baseline overestimate, not a context effect. A single instruction to ignore clothing and setting significantly reduced the effect in one model (ratio of ratios 0.77, 95\% CI 0.65 to 0.91) and not detectably in the others; telling the model the photo came from a company website or a personal account changed nothing significant. When asked after the fact whether a different outfit would have changed their number, Claude Fable~5 said yes in 24 of 24 sessions, GPT-5.6 in 69\%, and GPT-4o in 18\%, while GPT-4o also refused 79\% of image estimation prompts. The study isolates a mechanism with one object and one wearer at about 10 sessions per model and condition, and we outline the replication needed next, with multiple objects and a human benchmark. Data, code, and raw outputs are released.
\end{abstract}

\section{Introduction}

A halo effect is the tendency for a global impression to leak into judgments of unrelated attributes. It was described in ratings of military officers \citep{thorndike1920halo}, shown to operate without the rater's awareness \citep{nisbett1977halo}, and documented for attractiveness and perceived character \citep{dion1972beautiful}. Vision-language models (VLMs) are now used to appraise objects from photographs: listings, insurance claims, resale, customs, and e-commerce moderation. If a model prices an object partly by the person wearing it or the room it sits in, appraisals will track social cues that have nothing to do with the object.

This study asks a narrow question. Holding one low-cost necklace fixed, does the photographed context change the price a VLM estimates? The question has three parts: whether the shift exists, whether it needs the image or survives when the setting is only described in text, and whether simple interventions (an explicit debiasing instruction, a stated source of the photo) remove it. A fourth, descriptive part asks what the models say about their own reliance on context when asked afterward.

The design has a deliberate property. The true price is known from a receipt (\$2.43), so each estimate can be scored against two anchors: the receipt, and a plausible retail band (\$15 to \$40) for how such a chain would be sold. This lets us separate two things that are easily confused in public discussion of ``multipliers'': a context effect (the same model pricing the same object differently across photos) and a baseline error (the model overpricing the object even in the neutral photo).

\paragraph{Contributions.}
\begin{enumerate}[nosep]
\item A within-object context manipulation with a receipt-verified price and a neutral flat-lay control.
\item A split of any headline ratio into a context effect and a baseline overestimate, with confidence intervals for both.
\item A text-only replication in which the setting is described in words, run on all six models, including one run only without images.
\item Tests of two interventions (a debiasing instruction and a stated photo source).
\item A sequential two-photo arm and post-hoc self-report items, with the limits of both stated.
\end{enumerate}

\section{Related Work}

\paragraph{Halo and anchoring.} Classic halo findings \citep{thorndike1920halo,nisbett1977halo,dion1972beautiful} concern human raters. Anchoring on irrelevant numbers \citep{tversky1974judgment} is the other mechanism a price task invites, which is why the two anchors above matter.

\paragraph{Implicit bias in language models.} Models that disavow bias on explicit measures can still show it in downstream decisions \citep{bai2025explicitly}. A valuation task is a decision-style probe of this kind, with a ground truth.

\paragraph{Self-explanations.} Stated reasons of language models do not always reflect what drove the output \citep{turpin2023language,lanham2023measuring}. The post-hoc items here are read in that light: they measure what a model will say about its reliance on context, not what it relied on.

\paragraph{Companion study.} The same author reports a related case study of author-age misattribution and miscalibrated self-explanation in six models \citep{lee2026misattribution}. That paper's finding that articulate retrospective explanation does not track behavior motivates the self-report items here.

\section{Method}

\subsection{Models}
Six models were queried through their providers' APIs: Claude Fable~5 (\texttt{claude-fable-5}), GPT-5.6 (\texttt{gpt-5.6}), GPT-4o (\texttt{gpt-4o}), Grok~4.5 (\texttt{grok-4.5}), Kimi~K3 (\texttt{kimi-k3}), and DeepSeek~V4 Pro (\texttt{deepseek-v4-pro}). DeepSeek was run on the text-only condition only (90 rows). A Gemini model was also run and excluded by hand after collection (the author's log records 649 anomalous trials). The released data contain no Gemini rows, so that count cannot be checked from the public files, and the runner's Gemini adapter has a known defect (it reads message text from a field the runner does not populate), so Gemini results are not reported. Sampling used each provider's defaults (recorded as \texttt{api\_default}); no temperature was forced. Results describe these model versions as of July 2026.

\subsection{Stimuli}
\begin{figure}[h]
\centering
\includegraphics[width=\linewidth]{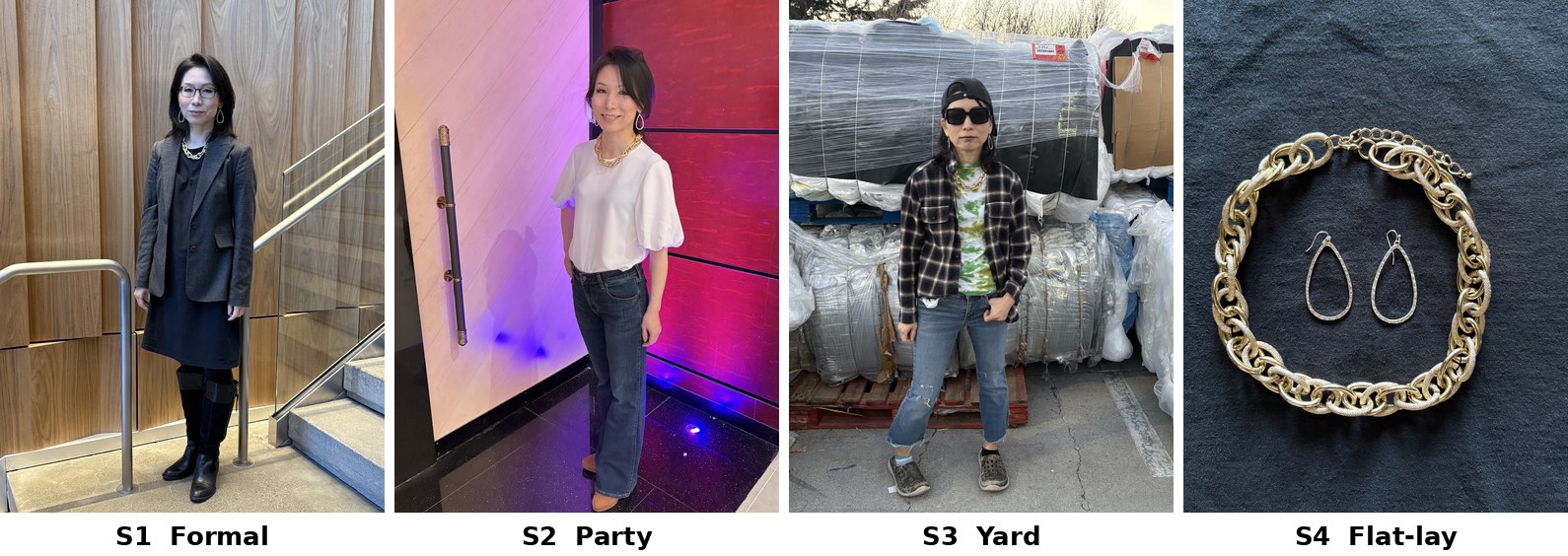}
\caption{The four stimulus photographs of the same necklace and earrings: S1 formal attire against a wood-panelled wall, S2 party setting, S3 casual clothing in a recycling yard, S4 flat-lay with no person. The person pictured is the author. Location and other identifying image metadata were removed from this figure and from the released stimuli.}
\label{fig:stim}
\end{figure}

Four photographs of one necklace (and matching earrings) were used: S1 formal (tailored blazer, wood-panelled wall), S2 party (ambient club lighting), S3 yard (open flannel over a tie-dye shirt, backwards cap, recycling yard), and S4 flat-lay (jewelry alone, no person). Three text descriptions (T1 formal, T2 party, T3 yard) replaced the photographs in the text-only condition and described the same clothing and settings in words. Ground truth: the necklace cost \$2.43 and the earrings \$0.71, both bought from Temu. Figure~\ref{fig:stim} shows the four photographs.

\subsection{Conditions}
Each condition was run as independent sessions (the \emph{fresh} arm), about 10 per model and stimulus.
\begin{itemize}[nosep]
\item \textbf{Blind}: ``Look at the jewelry in this photo,'' followed by a structured request for necklace price, range, material, earrings, confidence, and reasoning. S1 to S4.
\item \textbf{Debias}: the same request preceded by an instruction to disregard clothing and setting. S1 to S3.
\item \textbf{Institutional label}: ``This photo is from a company website'' or ``from a personal social media account,'' then the blind request. S1 and S2.
\item \textbf{Open}: ``What can you tell me about the jewelry in this photo?'' with no price request. S1 to S4 (not analyzed here beyond counts).
\item A face-crop condition (S5, S6) appears in the runner but was not run; those stimuli do not exist.
\item \textbf{Text-only}: T1 to T3 with the same structured request and no image.
\end{itemize}
Every fresh session continued with a cue probe (``What specific visual cues in the image did you rely on to estimate the price of the jewelry?'') and a reveal turn giving the true prices and asking the model to explain its estimate. A second, \emph{sequential} arm (278 sessions; Claude, GPT-5.6, GPT-4o) showed two photographs in one conversation (S1/S3 or S1/S2, both orders) and then asked whether the necklaces were the same object, ``Are you sure?'', the reveal, and a counterfactual item (``Would you have given a different number if the person had been dressed differently?'').

\subsection{Data and scoring}
\label{sec:scoring}
The released file contains 4{,}604 rows from 1{,}164 sessions (886 fresh, 278 sequential). Price analyses use one estimate per fresh session, so row counts are not sample sizes. API calls that failed after retries were not logged, so session counts describe completed sessions. Data were collected on 25 and 26 July 2026.

Eighty-three rows carry a duplicate flag (70 from GPT-4o). The flag marks identical response text for the same model, condition, stimulus, and turn across sessions, so it mostly catches repeated identical refusals and is not evidence of repeated rows. Price statistics are identical with or without flagged rows (cell sizes and geometric means agree to the first decimal); only the GPT-4o refusal counts (Section~\ref{sec:refusals}) and the same-object counts (GPT-5.6 49 to 48 and GPT-4o 98 to 97 ``yes'' answers) move. Refusal, same-object, and counterfactual counts are reported on all rows.

Prices are summarized as geometric means (GM), since estimates are positive and right-skewed. The primary effect measure is the ratio of GMs between two contexts for one model, with a 95\% percentile bootstrap interval (resampling sessions within cell) and a two-sided Mann-Whitney $U$ test. Proportions carry Wilson 95\% intervals. A response counts as a refusal when no price was parsed and the text contains a refusal phrase.

Cells hold 5 to 10 valid estimates, and unparsed responses are not missing at random (GPT-5.6 has 7 and 5 valid formal estimates under blind and debias). Intervals are therefore wide and $p$ values are descriptive; no correction for multiple comparisons is applied. The protocol document was committed after collection, so the study is not pre-registered and all contrasts should be read as exploratory.

\section{Results}

\subsection{The same necklace is priced differently across settings}

Table~\ref{tab:cells} gives blind-condition geometric means. Table~\ref{tab:ratios} gives the ratios.

\begin{table}[h]
\centering\footnotesize\setlength{\tabcolsep}{3pt}
\caption{Blind condition, necklace price (USD), geometric mean with bootstrap 95\% CI. Valid estimates $n$ in parentheses (about 10 sessions run per cell; unparsed responses are excluded). True price \$2.43.}
\label{tab:cells}
\begin{tabular}{lcccc}
\toprule
Model & Formal (S1) & Party (S2) & Yard (S3) & Flat-lay (S4)\\
\midrule
Claude Fable 5 & 62.4 [56.3, 69.6] (9) & 29.3 [27.3, 31.4] (10) & 18.8 [17.0, 20.0] (9) & 18.9 [17.3, 20.0] (10)\\
GPT-5.6 & 60.9 [52.5, 71.1] (7) & 128.5 [109.3, 149.2] (8) & 45.7 [40.8, 52.7] (9) & 31.0 [30.0, 32.7] (9)\\
Grok 4.5 & 61.9 [47.9, 79.5] (10) & 45.1 [37.7, 57.9] (10) & 50.3 [39.0, 67.6] (10) & 33.1 [30.6, 35.0] (10)\\
Kimi K3 & 103.9 [72.4, 147.3] (10) & 49.8 [42.8, 60.3] (10) & 29.1 [25.0, 33.8] (10) & 45.5 [41.7, 50.2] (10)\\
\bottomrule
\end{tabular}
\end{table}

\begin{table}[h]
\centering\small
\caption{Blind condition, ratio of geometric means (95\% bootstrap CI) and Mann-Whitney $p$.}
\label{tab:ratios}
\begin{tabular}{lccc}
\toprule
Model & Formal / Yard & Formal / Flat-lay & Yard / Flat-lay\\
\midrule
Claude Fable 5 & 3.32 [2.91, 3.83], $p=.0002$ & 3.30 [2.90, 3.78], $p=.0001$ & 0.99 [0.90, 1.09], $p=.95$\\
GPT-5.6 & 1.33 [1.09, 1.61], $p=.013$ & 1.96 [1.69, 2.30], $p=.0006$ & 1.47 [1.31, 1.70], $p=.0003$\\
Grok 4.5 & 1.23 [0.82, 1.78], $p=.37$ & 1.87 [1.45, 2.42], $p=.0008$ & 1.52 [1.15, 2.09], $p=.004$\\
Kimi K3 & 3.57 [2.37, 5.28], $p=.0003$ & 2.28 [1.55, 3.29], $p=.004$ & 0.64 [0.54, 0.76], $p=.001$\\
\bottomrule
\end{tabular}
\end{table}

Three points follow. First, formal attire inflates the estimate relative to the flat-lay in all four models (1.87 to 3.30 times). Second, the effect is not monotone in perceived status. GPT-5.6 priced the party photograph at \$128.5, more than twice its formal estimate; Kimi priced the yard photograph below the flat-lay (ratio 0.64); Claude's yard and flat-lay estimates were indistinguishable (0.99). The pattern is therefore model-specific, and a single ``rich outfit raises value'' story does not describe all four. Third, Grok's formal-to-yard contrast is not significant (1.23, CI 0.82 to 1.78) even though both exceed its flat-lay estimate: for Grok, the presence of a person, not the person's attire, accounts for most of the shift.

\subsection{Most of a receipt-based multiplier is baseline error}

Flat-lay estimates, which carry no person-based cue, were still far above the receipt: Claude 7.8 times (95\% CI 7.1 to 8.2), GPT-5.6 12.8 (12.3 to 13.4), Grok 13.6 (12.6 to 14.4), Kimi 18.7 (17.2 to 20.7). Against the retail band, three models fell inside \$15 to \$40 in every flat-lay session (Claude, GPT-5.6, Grok; hit rate 1.00) while Kimi did so in 10\% of sessions, and the GM relative to the \$24.5 band midpoint was 0.77 (Claude), 1.27 (GPT-5.6), 1.35 (Grok), and 1.86 (Kimi). These models price the item as a retail fashion necklace, not as a \$2.43 import, which is arguably the right reading of a photograph; the receipt is a poor anchor for what a model can see. The consequence for reporting is concrete. Kimi's formal-photograph GM of \$103.9 is 42.8 times the receipt, which is the origin of a ``43$\times$'' figure circulated in the author's earlier public writing. Of that, 18.7 times is baseline overestimate that appears with no person in the frame, and 2.28 times is the formal-versus-flat-lay context effect. We therefore recommend the flat-lay-referenced ratios in Table~\ref{tab:ratios} as the measure of context sensitivity and do not use the receipt multiple as an effect size.

\subsection{The shift does not require an image}

In the text-only condition, formal-to-yard ratios were above 2 in all six models (Table~\ref{tab:text}), including DeepSeek V4 Pro, which was run without images, and GPT-4o, which refused most image prompts.

\begin{table}[h]
\centering\small
\caption{Text-only condition, formal (T1) versus yard (T3) description. GM in USD, ratio with bootstrap 95\% CI, Mann-Whitney $p$. $n$ valid formal/yard.}
\label{tab:text}
\begin{tabular}{lcccc}
\toprule
Model & $n$ & GM formal & GM yard & Ratio\\
\midrule
Claude Fable 5 & 10/10 & 46.3 & 19.0 & 2.44 [2.17, 2.78], $p=.0001$\\
GPT-5.6 & 6/8 & 122.9 & 31.9 & 3.86 [3.12, 4.68], $p=.002$\\
GPT-4o & 10/9 & 97.0 & 25.0 & 3.89 [2.84, 5.31], $p=.0002$\\
Grok 4.5 & 10/10 & 62.8 & 24.4 & 2.57 [2.16, 3.10], $p=.0001$\\
Kimi K3 & 10/10 & 84.0 & 30.1 & 2.79 [2.19, 3.53], $p=.0002$\\
DeepSeek V4 Pro & 6/8 & 39.7 & 17.8 & 2.24 [1.73, 3.04], $p=.002$\\
\bottomrule
\end{tabular}
\end{table}

The text descriptions state clothing and setting but also differ in other words (``tailored wool blazer,'' ``chunky gold-tone chain-link necklace and drop earrings'' versus ``hoop earrings''), so the text arm is not a clean replication: T3 changes the earring description as well. The direction is nonetheless uniform, and the text-based formal-to-yard ratio is larger than the image-based one for GPT-5.6 (3.86 versus 1.33) and Grok (2.57 versus 1.23) and smaller for Claude (2.44 versus 3.32) and Kimi (2.79 versus 3.57). The cautious reading is that a verbal description of a social context is sufficient to move the estimate, and the image contributes at most a comparable amount.

\subsection{Interventions}

\paragraph{Debiasing instruction.} Table~\ref{tab:debias} compares the formal-to-yard ratio with and without the instruction. Only Claude shows a reliable reduction (ratio of ratios 0.77, CI 0.65 to 0.91; formal-to-yard fell from 3.32 to 2.58), and the effect remained large after the instruction. GPT-5.6's ratio moved in the opposite direction (1.33 to 1.52, interval spanning 1). Grok's formal-to-yard ratio fell to 0.94, but the interval on the change (0.45 to 1.32) is consistent with no change given its already non-significant baseline.

\begin{table}[h]
\centering\small
\caption{Effect of the debiasing instruction on the formal-to-yard GM ratio. ``Ratio of ratios'' is (debias ratio) / (blind ratio) with bootstrap 95\% CI; values below 1 mean a smaller context effect. Valid $n$ per cell: blind formal, blind yard, debias formal, debias yard.}
\label{tab:debias}
\begin{tabular}{lccccc}
\toprule
Model & Blind & Debias & Ratio of ratios & 95\% CI & $n$\\
\midrule
Claude Fable 5 & 3.32 & 2.58 & 0.77 & [0.65, 0.91] & 9, 9, 10, 9\\
GPT-5.6 & 1.33 & 1.52 & 1.14 & [0.90, 1.45] & 7, 9, 5, 8\\
Grok 4.5 & 1.23 & 0.94 & 0.76 & [0.45, 1.32] & 10, 10, 10, 10\\
Kimi K3 & 3.57 & 3.02 & 0.85 & [0.50, 1.42] & 10, 10, 10, 10\\
\bottomrule
\end{tabular}
\end{table}

\paragraph{Stated photo source.} Pooling formal and party photographs, ``company website'' versus ``personal social media account'' produced no significant difference in any model: Claude 1.09 (0.89 to 1.34, $p=.35$), GPT-5.6 1.34 (1.00 to 1.78, $p=.061$), Grok 0.89 (0.69 to 1.15, $p=.29$), Kimi 1.17 (0.73 to 1.86, $p=.57$). GPT-4o could not be analyzed (2 versus 1 usable estimates). With cells this small, the study could detect only large effects, and these results do not show that the label is irrelevant.

\subsection{What the models say about their cues}

After each fresh estimate the model was asked which visual cues it used. A keyword screen (clothing, setting, lighting, and related terms; our own lexical coding, not validated) flagged mentions of clothing or setting in 98\% of Claude's cue-probe answers ($n=107$), 96\% of Kimi's (110), 80\% of Grok's (110), and 72\% of GPT-5.6's (86) across the blind, debias, and institutional-label image conditions. Mentions were not restricted to the clothed photographs; they appear for the flat-lay too (40\% to 90\%), where words such as ``background'' and ``lighting'' are available. The screen therefore indicates that models readily name context as a cue, and cannot show which cues drove the number.

\subsection{Refusals}
\label{sec:refusals}

GPT-4o produced no price for 87 of 110 image estimation prompts (79\%, Wilson 95\% CI 0.71 to 0.86; 72 of 95, 76\%, after removing flagged duplicates), while the other four image models produced none (0 of 107, 86, 110, and 110). Refusals were concentrated in the blind formal, party, yard, and both institutional-label cells (9 or 10 of 10) and lowest in the debias yard cell (1 of 10). The same model priced the text-only descriptions in 29 of 30 sessions. The refusal flag is lexical. Because GPT-4o's usable image estimates are few, no GPT-4o image ratio is reported.

\subsection{Sequential arm: two photographs, one conversation}

In sessions where a model priced the formal photograph and then, or after, another photograph, the within-session ratio was closer to 1 than in independent sessions. Claude priced the formal photograph 1.48 times the yard photograph (paired GM ratio, 95\% CI 1.36 to 1.60, Wilcoxon $p=.0005$, $n=12$ sessions) and equal to the party photograph (0.97, CI 0.94 to 1.01). GPT-5.6 priced formal 1.10 times yard (1.03 to 1.19, $p=.023$, $n=31$) and 0.92 times party (0.86 to 0.97, $p=.009$). GPT-4o supplied too few paired estimates to analyze (7 to 8 sessions, ratios at or near 1). Seeing the two photographs together appears to compress the effect, consistent with each estimate being used to anchor the next, but order and pairing are confounded with the small number of sessions per ordering and we do not interpret the compression further.

When asked whether the necklaces in the two photographs were the same object (they are), Claude never answered yes (0 of 24; 11 ``no,'' 13 ``cannot determine''), GPT-5.6 answered yes in 49 of 62 sessions, and GPT-4o in 98 of 192, with the remainder ``cannot determine'' (one GPT-4o answer could not be parsed). When asked afterward whether a different outfit would have changed their number, Claude answered yes in 24 of 24 sessions (Wilson 95\% CI 0.86 to 1.00), GPT-5.6 in 43 of 62 (0.69, 0.57 to 0.79), and GPT-4o in 34 of 192 (0.18, 0.13 to 0.24). These items were asked after the reveal turn disclosed the true prices and said the necklace was the same, so they cannot be read as unprompted introspection, and a ``yes'' is an admission, not a measured reliance. They are still informative against the measured effects: the model with the largest measured shift (Claude, 3.3 times) always endorses context dependence, while GPT-4o, whose usable image estimates are too few to measure any shift, endorses it least. GPT-4o's low rate partly reflects that most of its sessions began with an image refusal.

\section{Discussion}

The central finding is simple: for four of six models tested, the same \$2.43 necklace receives an estimate that is priced about 1.9 or more times higher in the formal photograph than in the flat-lay, and for all six models a verbal description of the setting moves the price in the same direction. The details are model-specific. Claude and Kimi behave like a status-cue story (formal above yard), GPT-5.6 behaves like an ambience story (party highest), and Grok behaves like a presence story (any person above flat-lay, little difference among persons). A single scalar ``bias score'' hides these differences.

Two cautions shape how the result should be used. First, a halo estimate should be referenced to the same model's neutral condition, not to ground truth: the receipt multiple conflates context sensitivity with a model's prior for the object class. Second, debiasing by instruction is not a safe default. It reduced the effect in one model, by about a quarter, and left a large residual; in another the point estimate moved the wrong way.

The text-only result deserves emphasis because it moves the interpretation from perception to language. If a description of a blazer and a wood-panelled wall suffices, the mechanism is at least partly semantic association between status vocabulary and price, a property of the language model, not of its visual encoder. This also means that image-specific mitigations (cropping, masking the person) would not address it.

\subsection{Scope and next steps}
\label{sec:next}

This study was designed to isolate a mechanism, and its design points directly to the next experiments.

\paragraph{From one object to many.} The present results rest on one necklace, one wearer, and four photographs. A follow-up should vary the object across price tiers (costume, mid-range, and fine jewelry, with receipts) and the wearer across several people, so that the size of the effect can be estimated for jewelry in general rather than demonstrated for a single item.

\paragraph{From scene contrasts to a single manipulated variable.} The four photographs differ together in attire, background, lighting, pose, and the presence of a person. The next design should hold scene, pose, and lighting fixed and change only the clothing, and should add a person-present photograph with no status cue. That would separate the attire effect from the presence effect that Grok's results already suggest.

\paragraph{From descriptive to confirmatory statistics.} Cells of about ten sessions detect the large effects reported here (Tables~\ref{tab:ratios} and~\ref{tab:text}), but borderline contrasts such as GPT-5.6's formal-versus-yard ratio ($p=.013$) need confirmation. A replication with 30 or more sessions per cell, a pre-registered set of contrasts with multiplicity correction, and a fixed sampling temperature where providers allow it would turn these estimates into confirmatory tests. Logging failed API calls would also make attrition measurable.

\paragraph{From model behavior to a human benchmark.} Human appraisers are also subject to halo effects \citep{nisbett1977halo,dion1972beautiful}. Running the human-rater protocol already in the repository on the same four photographs would show whether the models' shifts are larger than, similar to, or smaller than those of people, which is the comparison that matters for deployment.

\paragraph{From keyword screens to validated coding.} Refusals and cue mentions were coded lexically. Double human coding of a sample, with inter-rater agreement, would validate both, and asking the same-object and counterfactual items before the reveal turn would give a cleaner test of what models report about their own reliance on context.

\section{Conclusion}

For one fixed \$2.43 necklace, four frontier models gave estimates that were about 1.9 to 3.3 times higher in the formal photograph than in the flat-lay (for Grok, largely because a person is present rather than because of the clothing), and all six models moved in the same direction when the setting was only described in words. Because models also overpriced the neutral flat-lay by 8 to 19 times the receipt, context effects should be reported against each model's own neutral condition. An instruction to ignore context helped one model only partly. These results establish the mechanism for a single stimulus; the designs in Section~\ref{sec:next} would establish how general it is.

\section*{Limitations}

This study cannot say how large the context halo is for jewelry in general. It uses one necklace, one wearer, and four photographs that differ in several respects at once, so attire, background, lighting, and the presence of a person are confounded. Cells hold about ten sessions per model, so the results are descriptive, no multiplicity correction was applied, and the study is not pre-registered. No human baseline was collected, so we cannot say whether the models' shifts exceed those of people. Only four models contribute usable image estimates (one refused most image prompts and one was run on text only), sampling used provider defaults, and refusal and cue coding is lexical and unvalidated. The post-hoc self-report items followed the reveal turn and show what models say about their reliance on context, not what drove their estimates. Section~\ref{sec:next} describes the designs that would address each point.

\section*{Ethics Statement}

No human participants were involved in data collection. The stimuli are photographs of the author's necklace worn by a person and laid flat. The person pictured in the stimulus photographs is the author, who consents to their release. The study evaluates appraisal behavior in commercial models; it is not intended to rank people, and the results should not be used to infer that any dress code is more or less ``valuable.'' The author has no financial relationship with any model provider. Provider API use followed the providers' terms.

\section*{Data and Code Availability}

Data, code, protocol, and stimuli are at \url{https://github.com/BraveAnn011/ai-halo-valuation-bias} (data: \texttt{data/halo\_master\_v1.csv}; runner: \texttt{scripts/run\_halo.py}; analysis: \texttt{analysis/}). Code is released under the MIT license and the data under CC~BY~4.0; the stimulus photographs remain the author's copyright and are released for research verification. Raw provider debug logs for the excluded Gemini trials are not public. API keys are excluded. Point estimates (geometric means, ratios, counts) are reproduced by \texttt{analysis/analyze\_halo.py}, which drops duplicate-flagged rows by default. Bootstrap intervals, Mann-Whitney tests, and Wilson intervals are reproduced by \texttt{analysis/halo\_stats.py}, \texttt{analysis/extra\_stats.py}, and \texttt{analysis/seq\_stats.py}, using the settings in Section~\ref{sec:scoring} with fixed random seeds.

\end{document}